\documentclass[journal]{IEEEtran}
\usepackage{graphicx}
\usepackage{cite}
\usepackage{picinpar}
\usepackage{amsmath}
\usepackage{amssymb}
\usepackage{url}
\usepackage{colortbl}
\usepackage{soul}
\usepackage{longtable}
\usepackage{microtype}
\usepackage{multirow}
\usepackage{pifont}
\usepackage{color}
\usepackage{alltt}
\usepackage[hidelinks]{hyperref}
\usepackage{enumerate}
\usepackage{siunitx}
\usepackage{xurl}
\usepackage{epstopdf}
\usepackage{pbox}
\usepackage{makecell}
\usepackage{amsfonts}
\usepackage{array,booktabs}
\usepackage{subfigure} 
\usepackage{bm}

\usepackage[ruled]{algorithm2e}
 
\begin{document}

\title{PSO-Based Optimal Coverage Path Planning for
Surface Defect Inspection of 3C Components with
a Robotic Line Scanner}

\author{Hongpeng Chen,~\IEEEmembership{Graduate Student Member,~IEEE,}
      Shengzeng Huo,~\IEEEmembership{Graduate Student Member,~IEEE,}
      Muhammad Muddassir,~\IEEEmembership{Member,~IEEE,}
      Hoi-Yin Lee,~\IEEEmembership{Graduate Student Member,~IEEE,}
      Anqing Duan,~\IEEEmembership{Member,~IEEE,}
      Pai~Zheng,~\IEEEmembership{Senior Member,~IEEE,}
      David Navarro-Alarcon,~\IEEEmembership{Senior Member,~IEEE}
        
\thanks{This work is supported by the grant from Shanghai Microintelligence
Technology Co. Ltd (No. P21-0078).}

\thanks{H. Chen, S. Huo, H. Lee, A. Duan, P. Zheng and D. Navarro-Alarcon are with the Faculty of Engineering, The Hong Kong Polytechnic University, Hong Kong, 999077, HKSAR (e-mail: hongpeng0925.chen@connect.polyu.hk, kyle-sz.huo@connect.polyu.hk, hoi-yin.lee@connect.polyu.hk, aduan@polyu.edu.hk, pai.zheng@polyu.edu.hk *Corresponding author phone: +852-2766-7816, email: dnavar@polyu.edu.hk).}
\thanks{ M. Muddassir is with the Faculty of Construction and Environment, The Hong Kong Polytechnic University, Hong Kong, 999077, HKSAR (e-mail: mmudda@polyu.edu.hk).	}
	}

\markboth{IEEE TRANSACTIONS ON INSTRUMENTATION AND MEASUREMENT}%
{Fan \MakeLowercase{\textit{et al.}}: }

\maketitle

\begin{abstract}
 The automatic inspection of surface defects is an essential task for quality control in the computers, communications, and consumer electronics (3C) industry. 
Traditional inspection mechanisms (viz. line-scan sensors) have a limited field of view, thus, prompting the necessity for a multifaceted robotic inspection system capable of comprehensive scanning. 
Optimally selecting the robot's viewpoints and planning a path is regarded as coverage path planning (CPP), a problem that enables inspecting the object's complete surface while reducing the scanning time and avoiding misdetection of defects.
However, the development of CPP strategies for robotic line scanners has not been sufficiently studied by current researchers.
To fill this gap in the literature, in this paper, we present a new approach for robotic line scanners to detect surface defects of 3C free-form objects automatically. 
Our proposed solution consists of generating a local path by a new hybrid region segmentation method and an adaptive planning algorithm to ensure the coverage of the complete object surface. 
An optimization method for the global path sequence is developed to maximize the scanning efficiency. 
To verify our proposed methodology, we conduct detailed simulation-based and experimental studies on various free-form workpieces, and compare its performance with two state-of-the-art solutions.
The reported results demonstrate the feasibility and effectiveness of our approach. 
\end{abstract}


\IEEEpeerreviewmaketitle

\section{Introduction}\label{sec:intro}

\IEEEPARstart{D}{efect} inspection is essential to quality control, process monitoring, and non-destructive testing (NDT) in the manufacturing industry~\cite{chen2022novel,chen2020arrival}. 
Specifically, manufacturing processes in the 3C industry are highly sophisticated and demand detailed and accurate defect inspection. 
Traditional defect inspection approaches typically rely on visual inspection of an intermediate/finished product by a quality control or quality check inspector. 
This sole dependence on human workers is a problem for regions and countries with a shortage of manpower~\cite{liu2021task, ming2020comprehensive, fu2024optimization}. Furthermore, human-based inspection is inherently subjective and, hence, prone to errors.
To address these issues, various researchers have reported the automatic surface inspection system for free-form components~\cite{li2022five, yang2023template}. 

Recently, automatic detection systems equipped with an industrial-grade line scanner, depth camera, and robotic manipulator have been developed to offer effective and rapid non-contact measurement~\cite{huo2021sensor, liu2022coverage}. 
During the defect inspection task, the robotic inspection system scans the surface of the target workpiece exhaustively from different viewpoints. Planning an inspection path can be considered as the coverage path planning (CPP) problem~\cite{molina2017detection}. 
Estimating a CPP strategy for automatic inspection consists of three tasks: (1) determining the viewpoints to measure the workpiece’s surfaces, (2) generating a sequence to visit all viewpoints in a time and kinematically optimal way, and (3) planning a feasible path to travel to each viewpoint. 
Additional criteria can be defined while planning the coverage path, including full coverage of the target surfaces and the resulting cycle-time for the inspection task~\cite{glorieux2020coverage}.
The existing CPP methods can be categorized into two coarse categories: two-dimensional and three-dimensional methods.

Various researchers reported two-dimensional (2D) CPP for mobile robots in floor cleaning, bridge crack monitoring, and weed mowing tasks~\cite{almadhoun2016survey,galceran2013survey } developed a motion planner for floor cleaning. 
Polyomino tiling theory was adapted to define reference coordinates and generate a navigation path to maximize the area coverage; Real-time experiments in different scenarios tested the planner on a Tetris-inspired shape-shifting robot. Hung M. La et al.~\cite{la2013mechatronic} proposed an autonomous robotic system for precise and efficient bridge deck inspection and identification, where a boustrophedon decomposition was applied to solve the CPP problem. 
Lim et al.~\cite{lim2014robotic} developed an automatic detection and mapping system for automatic bridge crack inspection and maintenance; They used an improved genetic algorithm to search for a CPP solution to minimize the number of turns and detection time while achieving an efficient bridge inspection. 
Danial Pour Arab et al.~\cite{pour2022complete} presented a CPP algorithm providing the optimal movements over an agricultural field. First, tree exploration was applied to find all potential solutions meeting predefined requirements, and then, a similarity comparison was proposed to find the best solution for minimizing overlaps, path length, and overall travel time. 

It must be remarked that 2D CPP methods cannot be adopted directly for a three-dimensional (3D) CPP problem, as the level of complexity in 3D space is much higher than in 2D space.
In most 2D applications, a complete planner map is available during planning. 
Most 3D CPP methods have to plan the paths from partial or occluded 3D maps.  
A CPP method for 3D reconstruction based on building information modeling used a robot arm and a lifting mechanism for wall painting at construction sites~\cite{zhou2022building}. 
It consists of a two‐stage coverage planning framework, a global planner that can optimally generate the waypoints sequence, and a local planner that can provide the mobile base pose.
The authors reported that this method could ensure coverage of all waypoints and improve painting efficiency. 
Hassan and Liu~\cite{hassan2019ppcpp} proposed an adaptive path planning approach cable of updating the paths when unexpected changes occur and still attain the coverage goal. 
Zbiss. K et al.~\cite{zbiss2022automatic} reported a path-planning method for collaborative robotic car painting. 
This proposed algorithm depends on computational geometry and convex optimization, and Morse cellular decomposition and boustrophedon algorithms are applied for path planning to generate a feasible and collision-free trajectory. 
A CPP method is based on Unmanned Aerial Vehicles (UAV) equipped LiDAR for bridge inspection~\cite{bolourian2020lidar}. 
This method combined a genetic algorithm and an A* algorithm to find a barrier-free and shortest path. This method planned the near-optimal and feasible path.

Recent studies on 3D CPP for industrial product quality detection focused on achieving full surface coverage of the workpiece with minimum inspection time are: 
Li et al.~\cite{li2018path} demonstrated a robust CPP method for aerospace structures based on their geometric features. Path planning relied on the feature graph construction through the Voronoi Diagram. Then, a search method is proposed to find this graph to decide the inspection sequence, and a convex hull-based approach is applied to avoid collisions.
Glorieux et al.~\cite{glorieux2020coverage} presented a targeted waypoint sampling strategy with the shortest inspection time for dimensional quality inspection of sheet metal parts. 
Liu et al.~\cite{liu2022coverage} developed an enhanced rapidly exploring random tree (RRT*) method and integrated the inspection errors and the optimal number of viewpoints into measurement cost evaluation for higher precision in quality inspection. 
Huo et al.~\cite{huo2021sensor} applied the nearest neighbor search algorithm to find a near-shortest scanning path aiming at convex free-form specular surface inspection.

Despite numerous recent developments, CPP for free-form surface inspection remains an open research issue. 
There are very few CPP solutions for line scanning robotic systems~\cite{kapetanovic2018side}. 
Compared with area-scan sensors, a line-scanning sensor is more suitable for defect inspection in industrial/manufacturing applications due to higher spatial resolution and lower production costs~\cite{steger2021camera, wang2022new}. 
Unlike a common area camera or other optical sensors that only work at some discrete positions, a line scanner utilizes only single beam scanning light to detect 3D objects when capturing images, and it needs to move continuously using a robotics manipulator along the coverage path. These features result in many traditional CPP methods being ineffective. 
Besides, existing CPP methods typically ignore the geometric differences among sole workpieces, negatively impacting detection results, particularly in the boundary of surfaces that could lead to unclear inspection.
Therefore, developing a novel CPP method for the automatic line scanning system becomes imperative and advantageous.

This paper aims to overcome the limitations of existing CPP methods for surface defect inspection. The challenge of providing minimal viewpoints for ensuring full coverage and precise defect inspection has been addressed in this work.
We focus on defect detection for free-form surfaces of 3C workpieces based on a robotic line scanning system.
This robotic system utilizes a 6-DOF robot manipulator with a line scanner to finish a full-coverage inspection path and a depth sensor to localize the workpiece. 
The proposed CPP method for robotics line scanning inspection consists of two parts: local path definition for accurate defect inspection and global time optimization for minimum scanning path. 
It incorporates the detailed requirements of 3C components surface inspection and the specific characteristics of a robotic line scanning system. Finally, the whole method can produce a full-coverage traversal path of all viewpoints efficiently. The inspection platform moves along this path to obtain surface defects.

These innovative contributions collectively provide a feasible and comprehensive method for solving
current quality monitoring challenges in industrial product
manufacturing, particularly in the 3C industry. To the best of authors' knowledge, this work first combines the optimization-based method, the segmentation-based method, and complex geometric representation for surface inspection according to a robotic line scanner. 

The main contribution of this paper includes: 
\begin{enumerate}[(1)]
    \item  A new region segmentation method and an adaptive region-of-interest (ROI) algorithm to define the local scanning paths for free-form surfaces with complex and high imaging quality assessment requirements.
    \item A Particle Swarm Optimization (PSO)-based global inspection path generation method to minimize the inspection time.  
    \item Detailed simulations, experiments, and comparisons to validate the proposed method. 
\end{enumerate}
The rest of this article is organized as follows. 
Section \ref{sec:ccp_for_inspection} describes the path planning problem for 3C component surface detection. 
Section \ref{sec:methods} presented the proposed CPP approach in detail. Section \ref{sec:results} shows the specific simulations, experiments, and comparisons on 3C components to validate the method's feasibility. 
Finally, Section \ref{sec:conclusion} concludes this article and discusses the limitations and future direction.

\section{Coverage path planning for inspection} \label{sec:ccp_for_inspection}

The CPP problem can be divided into two subproblems: 1) the local path definition is to generate view regions and partial scanning paths to meet the precise scanning and full coverage for 3C free-form workpieces. 2) global path planning aims to find an optimal or near-optimal sequence of all local paths~\cite{gerbino2016influence}. 

\begin{figure}[h]
	\centering
	\includegraphics[scale=0.3] {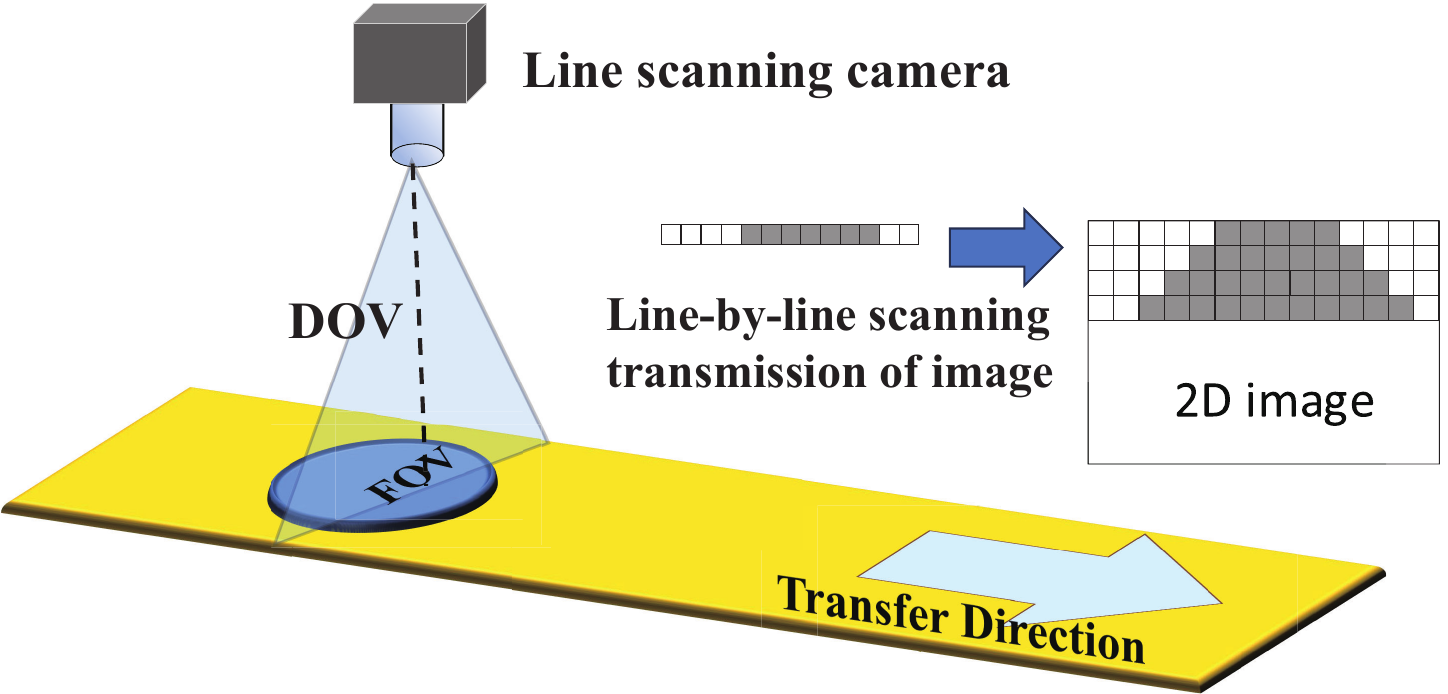}
	\caption{Conceptual representation of line scanning sensor}
	\label{fig:the basic priciple of line scanning}
\end{figure}

\begin{figure}[h]
	\centering
	\includegraphics[scale=0.5] {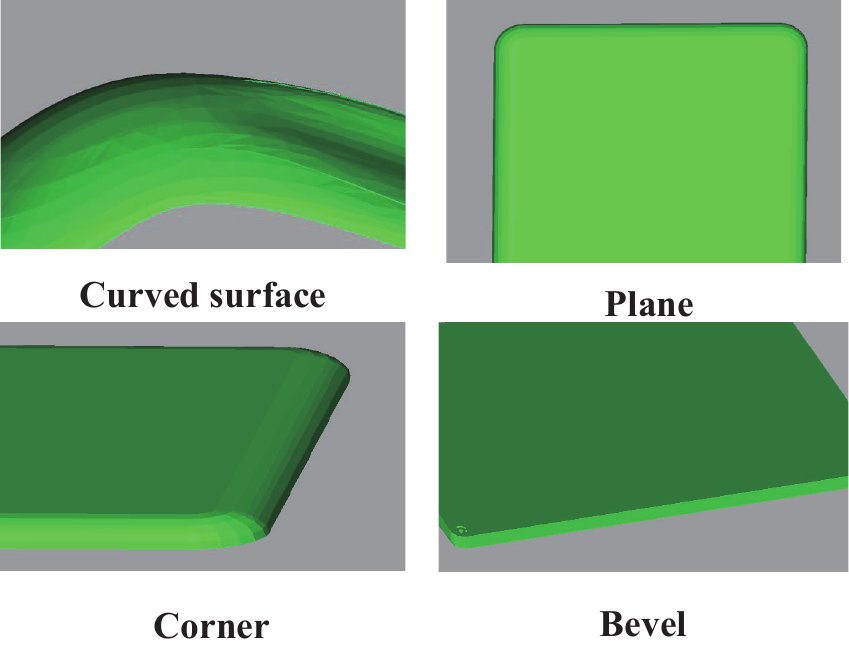}
	\caption{Geometric features of 3C workpieces}
	\label{fig:geometric feature}
\end{figure}

The key to the first sub-problem determines the position and orientation of each pair of viewpoints at both ends of local paths (the path between two consecutive viewpoints). The line-scan camera only captures an image line of pixels at a time, so the relative motion perpendicular to the line of pixels between the camera and object is necessary for 2D image acquisition during the defect inspection task (see Fig. \ref{fig:the basic priciple of line scanning}). In this automatic scanning system, the camera is moved with a robotics manipulator along the stationary object, and the direction of depth of view (DOV) of the camera should be perpendicular to the scanned region to ensure image quality. Therefore, the scanned area needs to be kept as flat as possible even if models of workpieces include many different geometric features (see Fig. \ref{fig:geometric feature}). In addition, each local path consists of two viewpoints at both ends, and the camera at the robotic end-effector could scan one viewpoint to another to inspect the surface defects of the regions corresponding to this local path. The change in the position of these two waypoints must be along one regular direction, whose orientations must remain as unchanged as possible to ensure the quality of acquired images. Besides, this sub-problem is also affected by some critical factors, such as field of view (FOV) and DOV~\cite{liu2022coverage}.

The global path planning problem involves finding the sequence and path connecting the selected viewpoints to minimize travel costs. This generated coverage path must reach all local paths with the shortest connection path. In other words, the objective is to find the minimum kinematic feasible path for the robot manipulator to target the scanning sensor at each viewpoint precisely through all local paths without colliding with any obstacles in the workspace.

This proposed method should provide a feasible coverage path that traverses all the local paths with minimum inspection time efficiently and automatically. Moreover, it needs to consider diverse measurement directions of local paths to ensure high detection precision. Generally, there are many local paths to evaluate the surface quality of the 3C components. To obtain precise defect original images, every scanning parameter is significant and could be set according to one new automatic method rather than the workers’ experience and opinion.


\section{Methodology}   \label{sec:methods}

\begin{figure*}[h]
	\centering
	\includegraphics[scale=0.38] {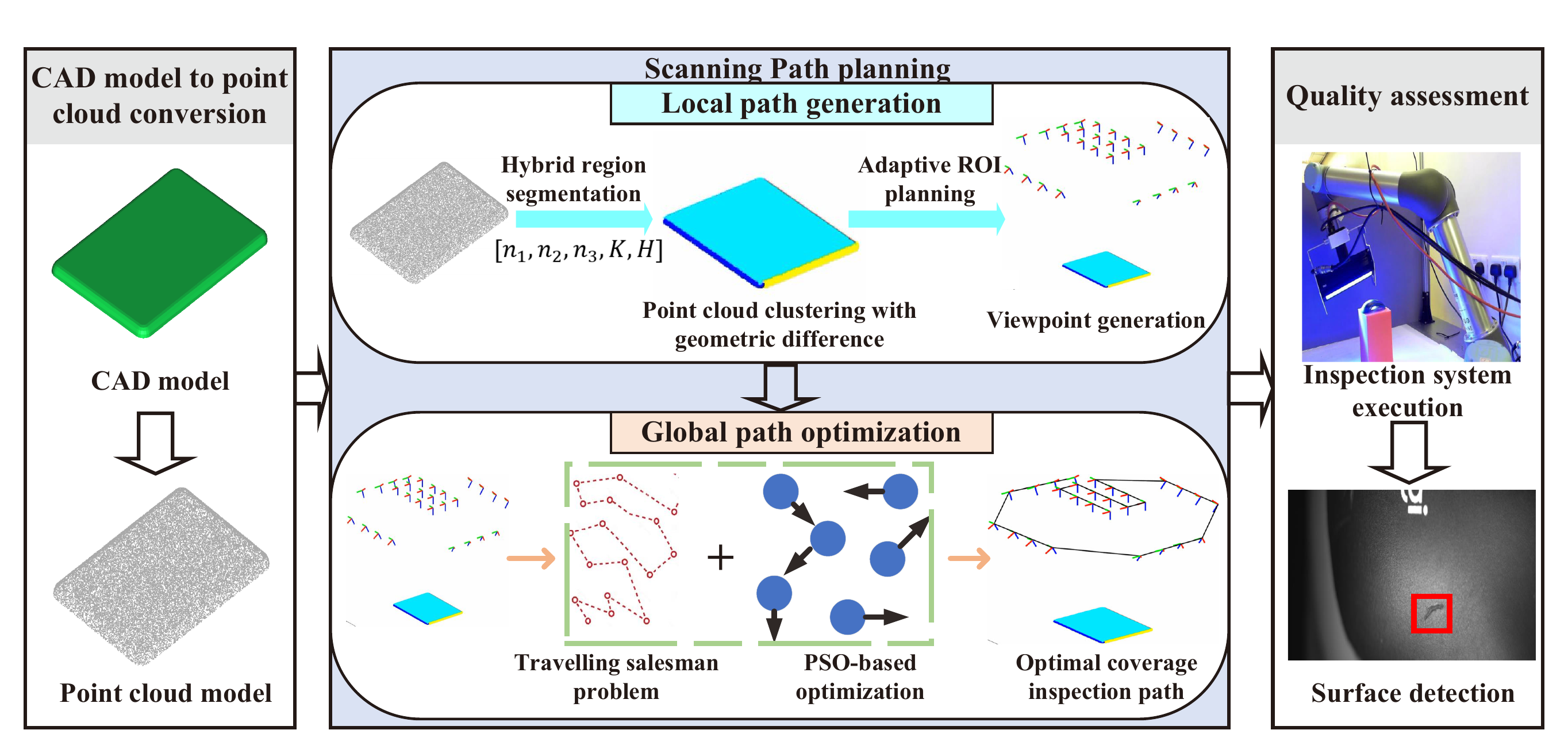}
	\caption{Framework of the proposed method}
	\label{fig:framework}
\end{figure*}

A CPP generation and optimization approach is presented based on the robotics line scanning system (see Fig. \ref{fig:framework}). This includes i) a new hybrid region segmentation method based on the random sample consensus (RANSAC) and K-means clustering method, ii) an adaptive ROI method to define the local scanning paths, and iii) one PSO-based global optimization approach for the minimum inspection time. This optimal path is then implemented for offline programming and surface detection, thereby improving the efficiency of the inspection of 3C components.

To extract the workpiece's geometry features, the 3D model is converted to a point cloud. The point cloud provides a detailed and accurate depiction of objects in 3D. Compared with mesh (STL) or other object descriptions, it facilitates the analysis of complex structures and key geometric information, including position, normal, and curvature~\cite{WOS:000627986900001}. The sampling procedure is based on selecting a series of points randomly and uniformly from the model to form a point cloud that can be used to segment and process all surfaces of the workpiece. The acquired point cloud $O$ consists of points $p_i=[x_i,y_i,z_i], i=1,2,..., m$ ($m$ is the total sampling number of $O$), which preserves the geometric information of all faces.

\subsection{Hybrid region segmentation based on RANSAC and K-means clustering}

The image acquisition characteristics of line-scan cameras necessitate the preservation of flat scanning areas to ensure optimal image quality. To accommodate this, it's essential to use a segmentation technique that can effectively divide a complex surface into flatter, more manageable segments. In our research, we introduce a novel hybrid region segmentation approach tailored for the unique contours of 3C component surfaces. This technique combines the robustness of RANSAC for identifying and modeling planes within the data with an enhanced version of the K-means clustering algorithm for precise segmentation, ensuring the division of the surface into suitable flat areas for high-quality image acquisition. 

Here, we use RANSAC to partition $O$ first. It includes two steps: producing an assumption by random samples and proving this assumption with the remaining data. Given different hypothesis geometrical models, RANSAC can identify planes, spheres, cylinders, and cones~\cite{xu2015investigation}. Since the flat regions are required for precise line scanning, RANSAC utilized the equation of a plane as a feature model in the proposed system. It selects $N$ sample points of $O$ and estimates the plane model parameters by those sample points. The position of a point is selected as an inlier if the distance between the point and plane is less than the fixed thresholds and the shape that contains the greatest number of outlier points could be split and extracted after multiple iterations. The plane model can be represented as
\begin{equation}\label{plane_model}
          aX+bY+cZ+d=0
  \end{equation}
  where $[a,b,c,d]^T$ is the plane model parameter, and $[X,Y,Z]^T$ denotes any point in the 3D coordinates. 

  This method can extract a nearly planar point cloud region $C_0$ when the best plane model has been identified. RANSAC does not require complex optimization or high memory resources so that we can obtain $C_0$ rapidly. However, the remaining point cloud $O^r$ with the size $\eta^r$  cannot be segmented clearly by this approach since $O^r$ consists of bevels, curved surfaces, and other complex geometrical information. 

The traditional K-means clustering methods regarded the region segmentation as a clustering analysis problem of surface geometric features. They applied the position and surface normals of the point cloud for segmentation, which are not appropriate for workpieces with significant variations in curvature or many bevels and corners~\cite{li2018leaf, liu2020method}. Therefore, some different factors should be considered to describe the features of the object. The enhanced K-means clustering is proposed in this paper to process $O^r$. In the standard K-means method, the number of clusters $N$ dramatically affects the performance of this method, and many trials are required to find a near-optimal $N$ in some classical methods~\cite{WOS:000290138700014}.  In this developed method, we apply not only the corresponding surface normals $n_i^r=[n_{ix}^r,n_{iy}^r,n_{iz}^r]$ of the points in $O^r$ but also the Gaussian curvature $K_i^r$ and Mean curvature $H_i^r$ of each point $p_i^r$ in $O^r$ as the inputs of the enhance K-means clustering. Besides, a feasible weighting factor $\omega$ among $n_i^r$, $K_i^r$, and $ H_i^r$ is determined through many manual experiments. $K_i^r$ is the product of the principal curvatures of $p_i^r$, and it neutralizes the maximum and minimum curvatures.  A positive Gaussian curvature value means the surface is locally either a summit or a valley, while a negative value illustrates the surface locally consists of saddle points. And zero Gaussian curvature indicates the surface is flat in at least one direction, like a plane or cylinder~\cite{li2019automated}. In mathematics, the mean curvature of a surface presents the curvature of an inset surface in Euclidean space or other ambient spaces. The curvature of the point can be represented by $c_i^r=[K_i^r, H_i^r]$. With adding these two parameters in this enhanced K-means method, the clustering quality can be improved than before, so the geometric feature of the point of $O^r$ is presented as $I_i^r = [n_i^r,c_i^r]$. Besides, we present a method to automatically adjust $N$ since $N$ affects the result of the classification, and the traditional techniques set one fixed $N$, whose drawback is its poor flexibility. The algorithm depends on a two-looped 1D search, with the inner loop for similarity comparison and the outer loop for iterating $N$. The iteration can end when the largest intra-class difference is smaller than a threshold $T$. The entire procedure of this enhanced K-means method is illustrated in Algorithm \ref{method1}. 

   For the outer loop, we represent the feature vectors of the $N$-cluster set as 
     \begin{equation}\label{feature vector}
          Q_j=[q_n,q_c] \quad
          q_n=[q_1,q_2,q_3] \quad
          q_c=[q_4,q_5]       
  \end{equation}
  
  $Q_j$ is one 5-dimensional vector ($j=1,2...,N$). All of them can be initialized with a random value. Afterward, the procedure goes into the inner loop, composed of two steps: 1) similarity comparison and 2) updating. In the first step, cosine similarity is used in this proposed method for assessing the similarity between $I_i^r$ and $Q_j$, which is considered as a measure of similarity between two sequences of numbers in data analysis~\cite{kiricsci2022new}. The similarity $\alpha _{ij}$ is described in detail as follows:
     \begin{equation}\label{simlarity comparision}
          \alpha _{ij} =\omega _1\cos \left(\frac{n_i^r \cdot q_n }{\left | n_i^r  \right | \cdot \left |  q_n\right | } \right)+\omega _2\cos \left(\frac{c_i^r \cdot q_c }{\left | c_i^r  \right | \cdot \left |  q_c\right | } \right)
  \end{equation}
 where $\omega _1$ and $\omega _2$ are the weighting factors for $\alpha _{ij}$, and they are set as 0.6 and 0.4 respectively in this method based on many trials. 
 
 Then, this method should find the cluster $C_j$ with the smallest $\alpha _{ij}$ and extract the corresponding $p_i^r$ and $I_i^r$ to it. The next step is to determine whether the classification has met the termination condition. For each cluster $C_j$, the termination parameter $\lambda _j$ is calculated from the maximum intra-class difference $D_j$ as:
   \begin{equation}\label{termination parameter}
     \lambda _j=\begin{cases}
     0, D_j>T\\1,else
     \end{cases};
     D_j= \max_{i}  \alpha _{ij}
  \end{equation}
  
    $\beta _t$ represents the sum of $ \lambda _j$ from every region $C_j$ at this iteration $t$ .If $\beta _t$ = $N$, the current segmentation is satisfactory and the algorithm can finish iteration. Otherwise, the procedure continues. In this stage, the search direction should be considered since the method includes two loops, the inner one that compares similarity and clusters concerning $N$ and the outer one that increases the value of
$N$ gradually. The change relies on the performance of $\beta _t$. If the performance deteriorates at the iteration step $t$ (i.e. $\beta _t$ is smaller than $\beta _{t-1}$),
the inner loop must stop immediately and a new outer loop starts with $N{\leftarrow}N+1$ because the current $N$ is not ideal. If the performance is better(i.e. $\beta _t$ is larger than $\beta _{t-1}$), the search within the inner loop continues.

Before switching to the next inner iteration, all feature vector $Q_j=[q_n,q_c]$ are updated to improve the representation level:
\begin{equation}\label{termination parameter1}
     q_n= \frac{\frac{1}{ \eta _j} \sum_{i=1}^{\eta _j} n_{ij}}{ \left \|\frac{1}{ \eta _j} \sum_{i=1}^{\eta _j} n_{ij}  \right \|}  \quad\quad
     q_c= \frac{\frac{1}{ \eta _j} \sum_{i=1}^{\eta _j} c_{ij}}{ \left \|\frac{1}{ \eta _j} \sum_{i=1}^{\eta _j} c_{ij}  \right \|}
\end{equation}

  where $n_{ij}$, $c_{ij}$ and $\eta_j$ are $i$-th normal feature vector in $C_j$, curvature feature vector in $C_j$ and the size of the $C_j$ separately.
  
  The proposed algorithm only takes the limited features of the region $C_j$ into consideration, which can lead to a high sparsity of the clustered points within the same region. Therefore, Euclidean cluster extraction is implemented as a post-processing step to verify if it is necessary to subdivide the region $C_j$ into two new regions according to the location of the points in it.

  The similarity threshold $T$ should be selected before region segmentation. 
If $T$ is large, the segmentation process needs more computation time to cluster the point cloud, which could reduce the overall clustering efficiency. 
On the contrary, a smaller value of $T$ groups the different features into the same cluster $C_j$, which degrades the segmentation accuracy. 
Consequently, selecting this component must balance the segmentation accuracy and calculation efficiency. 
0.64 is an optimal value for $T$, found by hit and trials.

\begin{algorithm} \label{method1}
      \caption{The enhanced K-means Region Segmentation}
      \KwIn {$T, O_r$}
      \KwOut {$C_j,j=1,2...,N$}
       \While {$\beta_t<N$}  {
       Initialize $Q_j$ randomly, $ j=1,2...,N$  \;
       \While {$\beta _t \geqslant \beta _{t+1}$} {
        \For{$i=1:\eta^r$}{
        Compute similarity $\alpha _{ij} {\leftarrow} (3) $\;
        $ j_{argmin} {\leftarrow}  \mathop{\arg\min} \alpha _{ij},$ take $ p_i^r{\rightarrow} C_j  $ \;
        }
        Calculate:    $\beta _t{\leftarrow} (4)$ \;
        Update: $Q_j {\leftarrow} (5) $\;
       }
       $N {\leftarrow} N+1 $ \;
    }
\end{algorithm}

\subsection{Adaptive ROI Based Path Planning}

The local paths are generated according to the proposed planning method, which takes the segmented region $C_j$ as input. Due to the synchronization of the scanning inspection of the line camera and the robot's motion, every viewpoint in these local paths should be produced through a feasible method for accurate detection, and all local paths are required to cover the whole region $C_j$ of the workpiece. Hence, this part presents an adaptive ROI method for generating local paths that aim to adapt scan paths and viewpoints to the various shapes of objects

\begin{figure}[h]
	\centering
	\includegraphics[scale=1] {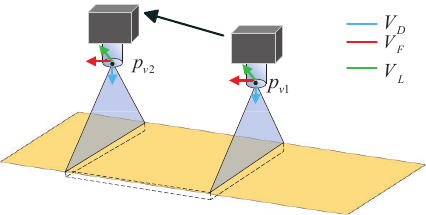}
	\caption{Cuboid coverage generation of line scanning camera during linear motion}
	\label{fig:Cuboid coverage generation}
\end{figure}

 Given that the scanning sensor acquires images in horizontal lines, the area it covers during linear movement can be visualized as a three-dimensional rectangular prism. This space encompasses the DOV represented by $V_D$, the FOV depicted by $V_F$, and the vector of linear motion $V_L$ (see Fig.~\ref{fig:Cuboid coverage generation}). Besides, the key of this approach is to determine the position $\mu =[x,y,z]$ and pose $i=[\vec{d},\vec{l}]$ of the viewpoints ($v^{p},v^{p*}$) at both ends of a local path $G_t,t=1,2,...,U$. The pose $i$ is described by the direction $\vec{d}$ of $V_D$ and the direction $\vec{l}$ of $V_L$.
 
To make the geometric scanning model effective and keep the detection accuracy of this system, our algorithm further segments every $C_j$ into 3 sub-regions $W_{jf}, f=1,2,3$. Due to the irregular shape of each $C_j$, we stipulate that the $C_j$ is divided into 3 sub-region $W_{jf}$ evenly following the direction $\vec{k} $ of the longest length of each $C_j$ and the scanning motion is also along $\vec{k} $ for every area ($\vec{l}=\vec{k}$). In addition, we define that $\vec{d}$ is the reverse direction of the surface normal $\vec{w_{jf}}$ of this $W_{jf}$ ($\vec{d}=-\vec{w_{jf}}$). 

Thus, the corresponding $\mu_1,\mu_2$  are located on :
  
\begin{equation}\label{ROI_1}
      \mu  = \tau-\vec{w_{jf}}\cdot |V_D|
  \end{equation}
  
 The center of the sub-region $W_{jf}$  is regarded as $c_{jf}=[c_x,c_y,c_z]$, and the intersections $ \tau_1,\tau_2$  of the $W_{jf}$ 's edge and the line $\vec{k} \cdot c_{jf}$ are deemed as the inspection points of viewpoints $v^{p},v^{p*} $  at both ends of a local path $G_t$ on this sub-region surface. $|V_D|$ is the magnitude of $V_D$.
    \begin{figure}[h]
	\centering
	\includegraphics[scale=0.4] {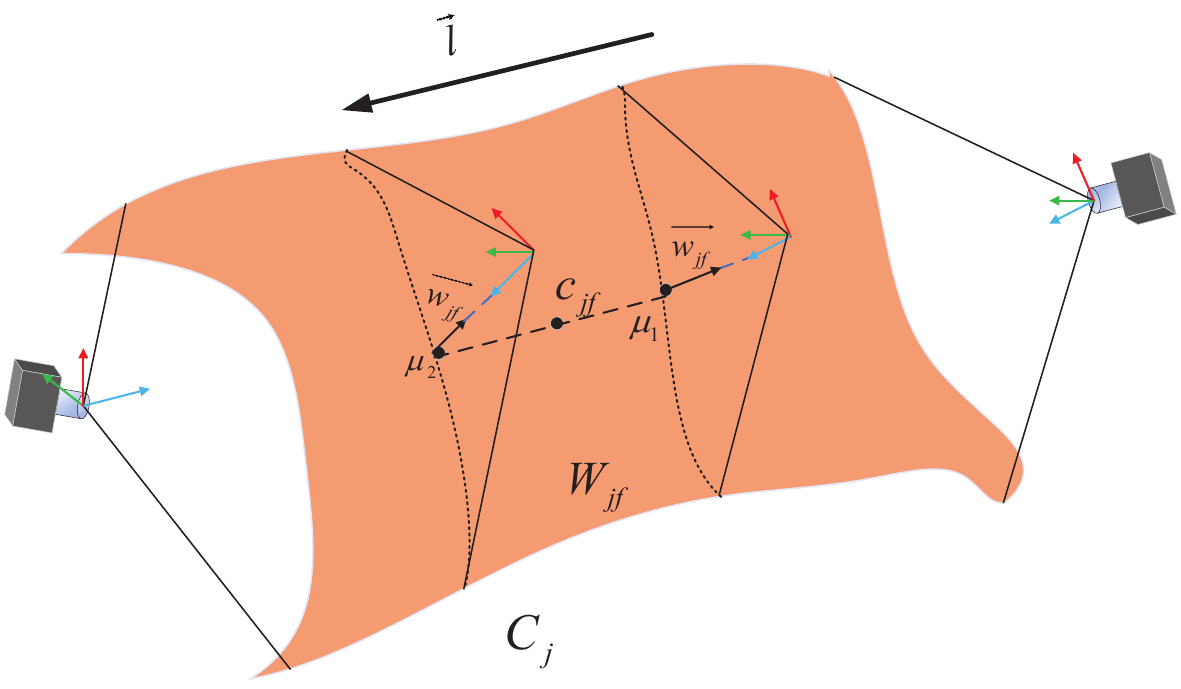}
	\caption{Further region segmentation and linear path planning}
	\label{fig:ROI}
\end{figure}

 \subsection{PSO-based global path optimization}
  Based on the local path definition in the previous step, the next challenge is to sequence these paths optimally to scan the entire surface of the free-form workpiece efficiently. The goal is to minimize the total movement time of the robot, assuming the sensor moves at a constant speed during the inspection. Practical demands dictate that the robotic arm must pass through all predetermined viewpoints to accomplish the scanning inspection. This sequencing conundrum is akin to the Traveling Salesman Problem (TSP), where we seek the most time-efficient route. The necessity of solving this problem lies in the efficiency gains for the inspection process, ensuring the robotic manipulator operates in the most time-effective manner without compromising the inspection quality~\cite{claro2023energy}. The TSP is one integrated optimization problem and nondeterministic polynomial time (NP)-hard. The problem of global path planning can be formulated 
  
  \begin{equation}\label{TSP}
     \min\left \{ \sum_{t=1}^{U} \sum_{s=1}^{U-1} T_t^{scanning}+T_s^{across} \right \} 
  \end{equation}
 where $T_t^{scanning}$ is the cost time of passing every local path $G_t$, $T_s^{across}$ means the cost time from $G_t$ to $G_{t+1}$ and $U$ represents the total number of local paths. The cost time in the context of the robot manipulator's end-effector is determined by the straight-line distance between two viewpoints, considering the constant speed of movement. In contrast to the general TSP, our scenario requires sequential traversal of adjacent viewpoints within the same local path to ensure optimal inspection performance. This constraint is imposed due to the limitations of region segmentation and the necessity for adaptive ROI local path definition. The limitation can be summarized as
    \begin{eqnarray}\label{TSP_lit}
     T_t^{scanning}(G_t)=\begin{cases}
      T(v_t^p\rightarrow v_t^{p*})\\ T(v_t^{p*}\rightarrow v_t^p)\\
     \end{cases}\\    T_s^across(G_t,G_{t+1})=\begin{cases}
     T(v_{t}^p\rightarrow v_{t+1}^{p})\\T(v_{t}^p\rightarrow v_{t+1}^{p*})\\T(v_{t}^{p*}\rightarrow v_{t+1}^{p*})\\T(v_{t}^{p*}\rightarrow v_{t+1}^{p})
     \end{cases}
  \end{eqnarray}
   The prior studies on this problem include branch and bound linear programming, and dynamic programming methods~\cite{shang2020co, xu2022path}. However, as the quantity of targets expands, the computation of a feasible path becomes exponentially more difficult, and obtaining the global optimal solution becomes more challenging.  In the proposed method, the PSO-based method is employed to tackle the TSP due to its inherent adaptability in finding solutions to this issue. After selecting the shortest path, the optimal general path sequence can be acquired in this step.
In PSO~\cite{karim2021hovering}, a swarm of particles are used to describe the possible solutions. Every particle $\xi$ is related to two vectors in $D$-dimension space, i.e.,
the velocity vector $\bm{V_{\xi}}=[V_{\xi}^1,V_{\xi}^2,...,V_{\xi}^D]$ and the position vector $\bm{X_{\xi}}=[X_{\xi}^1,X_{\xi}^2,...,X_{\xi}^D]$. Both of them are initialized by random vectors. During the PSO process, the velocity and position of particle $\xi$ on dimension $d$ are updated as~\cite{zhan2009adaptive}:

\begin{align}
    V_{\xi}^{d}={}&  \omega V_{\xi}^{d}+c_{1} \operatorname{rand}_{1}^{d}\left(pBest_{\xi}-X_{\xi}^{d}\right)  \notag\\
    &+ c_{2} \operatorname{rand}_{2}^{d}\left(gBest-X_{\xi}^{d}\right)
    \\
    X_{\xi}^{d}={}&X_{\xi}^{d}+V_{\xi}^{d}
\end{align}

where $ \omega$ represents the inertia weight (it is set as 1), and $c_1$ and $c_2$ are random numbers within [0,1]. $pBest_{\xi}$ is the position with the best fitness value for the $\xi$th particle and $gBest$ is the best position in the global.
The main steps of PSO are:
\begin{enumerate}[(1)]
    \item Initialize all particles, including their velocity and position. 
    \item Establish the fitness function and calculate the fitness value of each particle, \item Update the $pBest_{\xi}$ and $gBest$.
    \item Update the velocity and position of each particle according to (10) and (11).
    \item Increase the number of iterations, Go to step 3 and repeat until the termination condition.
\end{enumerate}

\subsection{Suitability of the proposed method}
Due to the uniqueness of the line-scan camera, it is necessary to propose a new CPP method for a robotics line scanner. In our method, a new hybrid region segmentation method can provide precise local scanning paths.  It is hard to obtain accurate region segmentation results using only the RANSAC method or k-means clustering. The RANSAC method can detect a region with planar geometry. It can also remove some points with minimum curvature from the entire point cloud, enhancing the computation speed of the whole procedure~\cite{su2022building}. Furthermore, it can effectively remove outliers, thereby improving the accuracy of the subsequent K-means clustering process. 
The conventional K-means clustering methods used the location and surface normals of the point cloud for region segmentation, which are not valuable for objects with large variations in curvature or some corners. In the enhanced k-means method, we do not only use the surface normals but also Gaussian curvature and Mean curvature, both of which can evaluate the geometric features comprehensively.
For the global path sequence optimization, the PSO-based method is used to solve it. However, the constraints and requirements are different from the previous work due to the limitations of region segmentation and the necessity for local path definition.

As compared to state-of-the-art region segmentation methods, such
as deep learning, the hybrid region segmentation method does not require complex data training before this procedure. Through direct processing for geometric features rather than certain regression feature analysis, it is time-efficient to attain accurate segmentation. This proposed method is more suitable for real-time or low-cost industry situations

\section{Case Study}   \label{sec:results}

To illustrate the performance of the proposed method, we provide two case studies for simulation tests (Case 1: a camera lens, Case 2: a Computer fan) and two case studies for experimental evaluation (Case 3: a tablet back cover, Case 4: upper part of computer mouse) on 3C component surface inspection. A state-of-the-art CPP method is also used for comparison with the developed method in Section \ref{ssec:case_study}.

\subsection{Case study setup}
\begin{figure}[h]
	\centering
	\includegraphics[width=\columnwidth] {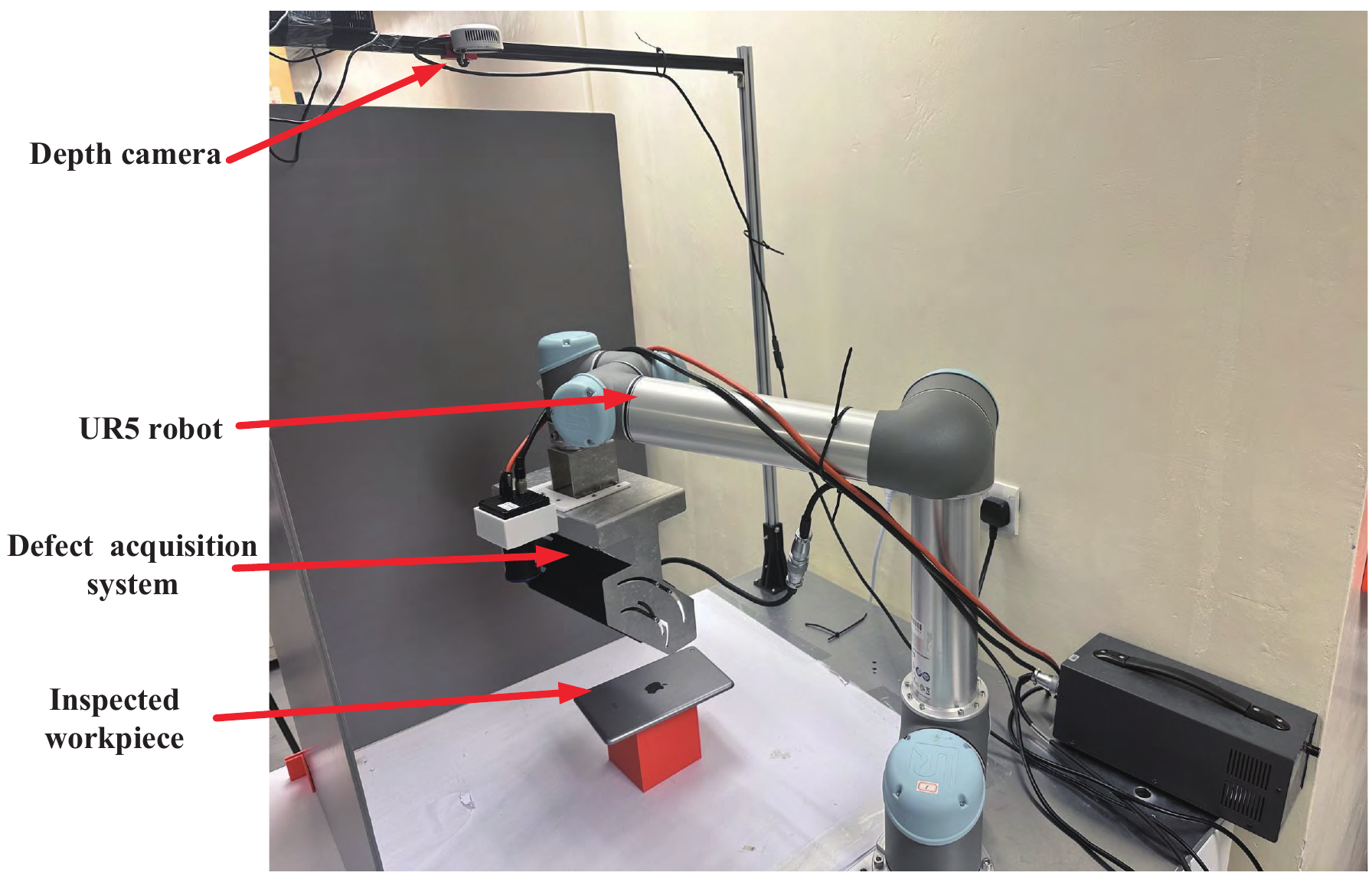}
	\caption{Experimental setup of the automated line scanning system }
	\label{fig: experiment platform}
\end{figure}

Fig. \ref{fig: experiment platform} shows the experimental setup for evaluating the proposed method. 
A custom-made end-effector housed the defect inspection system consisting of a line scanning sensor (Hikvision MV-CL041-70GM camera) and a uniform line illumination source (TSD-LSH230200-B from TSD company).
The Intel RealSense L515 LiDAR camera was mounted on the top of the workspace to capture the real-time stream of point clouds.
The pose of the workpiece was estimated using the point clouds from LiDAR.
An analog control box with a high-power strobe ensures an adjustable and stable voltage for the light source. 
The system consisted of a UR5 manipulator from Universal Robots to manipulate the end-effector in order to scan the workpiece automatically. 
The entire automated line scanning framework is based on ROSon Linux PC, which can simultaneously monitor the sensors (line scanner, depth sensor) and control the actuator (manipulator).

    \begin{table}[h]
\centering
\caption{The parameters used in the case study}
\begin{tabular}{cccccc}

   \toprule
   Parameter &  Value  \\
   \midrule
    $V_D$(mm) & 300    \\
    $V_F$(mm) & 70     \\
    Sampling frequency(Hz)  & 10000\\
    Image resolution & 3000*680\\
   \bottomrule
\end{tabular}
\label{tab:parameter}
\end{table}

The line velocity and acceleration of the manipulator's end-effector were empirically set to 0.05 $m/s$ and 0.5$m/s^2$, respectively. 
During trajectory execution, the robot manipulator followed a constant line speed to maintain consistency of image acquisition (the acquisition line rate of the scanner is 3000 $line/s$).  
Table~\ref{tab:parameter} summarizes the other parameters for the line scanning system used for the experiment. 

\subsection{Path generation and defect inspection}
\begin{figure}[h]
	\centering
	\includegraphics[scale=0.22] {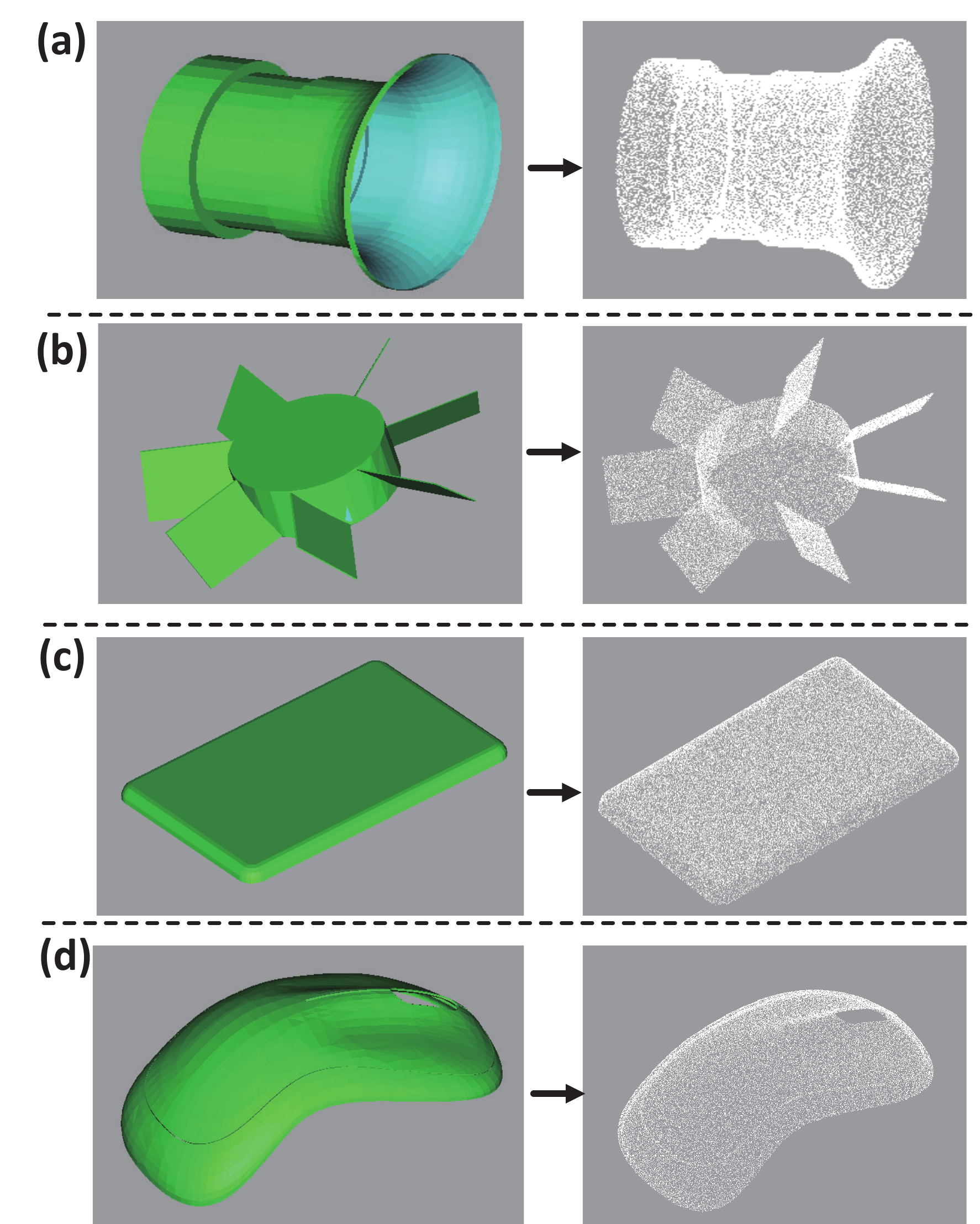}
	\caption{The model and point cloud of workpieces, (a) Case 1: the camera lens; (b) Case 2: the computer fan; (c) Case 3: the tablet back cover; (d) Case 4: upper part of computer mouse }
	\label{fig: the model of workpiece}
\end{figure}

Fig. \ref{fig: the model of workpiece} presents four 3C component models. 
Each 3D mesh model (or CAD model) was converted into a point cloud to identify the geometrical features through uniform and random sampling~\cite{WOS:000237574800012}, as shown in Fig. \ref{fig: the model of workpiece}.
Some geometrical features, such as surface normals, Gaussian curvature, and mean curvature, are computed by a point cloud processing software named CloudCompare~\cite{10081460}. 
Then, the point cloud was inputted into the proposed method for estimating the scanning path. 

\begin{figure}[h]
	\centering
	\includegraphics[scale=0.26] {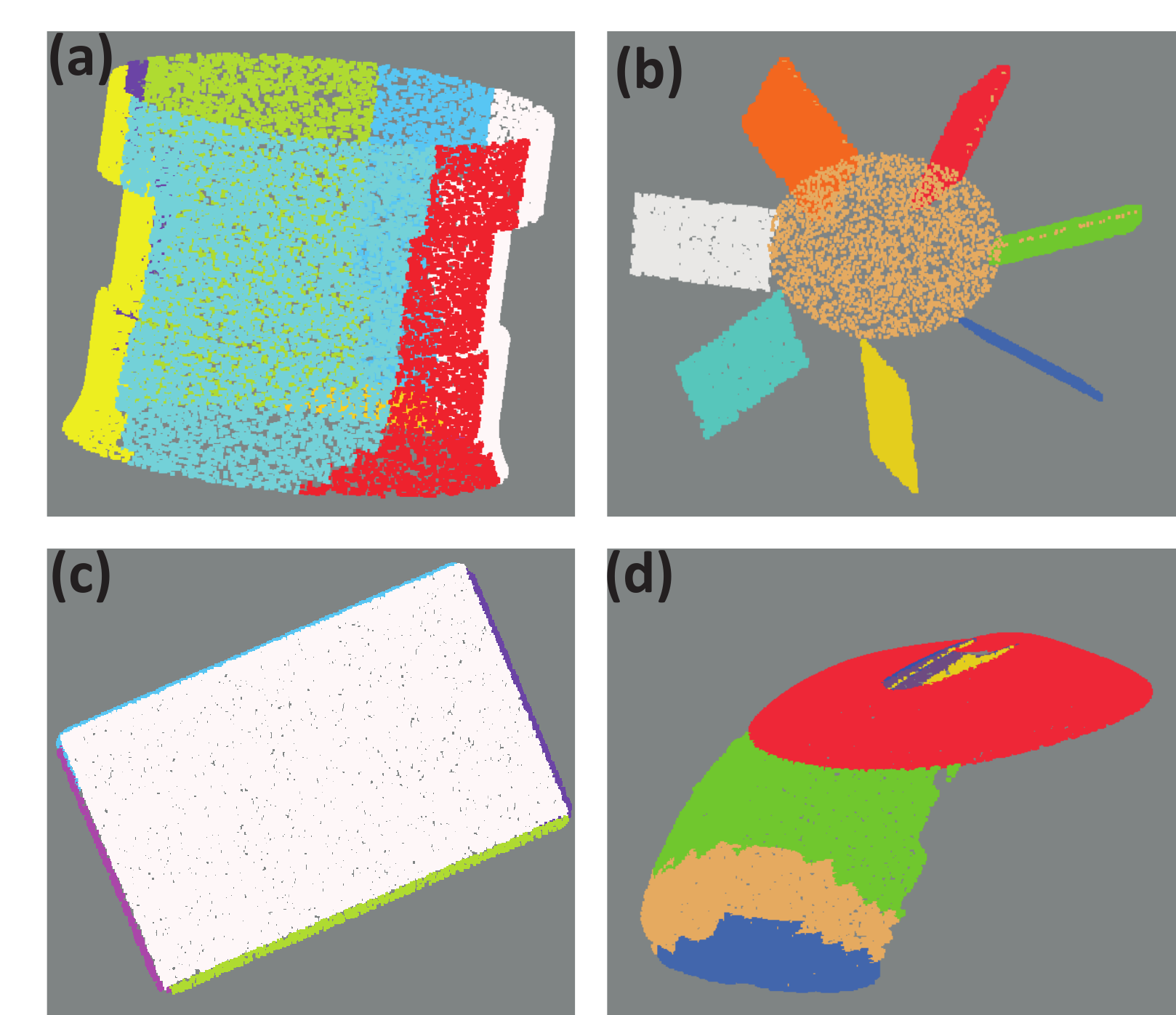}
	\caption{The segmentation result by the proposed method, (a) Case 1: the camera lens; (b) Case 2: the computer fan; (c) Case 3: the tablet back cover; (d) Case 4: upper part of computer mouse }
	\label{fig: The segmenation result by the proposed method}
\end{figure}

The results from the hybrid segmentation method are shown in Fig. \ref{fig: The segmenation result by the proposed method}, where the different colors indicate various segmented regions (or clusters). 
Here, the methods used RANSAC to cluster the plane region. 
In Case 3 and Case 4, a significant portion of the planar/near-planar region has been grouped in one cluster, as shown in Fig. \ref{fig: The segmenation result by the proposed method}(c). 
Initial clustering using RANSAC significantly reduces the processing time. 
After the hybrid unsupervised region segmentation, the surfaces with similar geometric features were clustered together. 
Fig. \ref{fig: The segmenation result by the proposed method} shows the four geometrically diverse workpieces, and each is divided into different regions based on the features. 
Some segmentation errors will remain due to the uncertain nature of computing features, but if they do not affect the scanning path generation. 

\begin{figure}[h]
    \centering
    \begin{subfigure}
        \centering
        \includegraphics[scale=0.3]{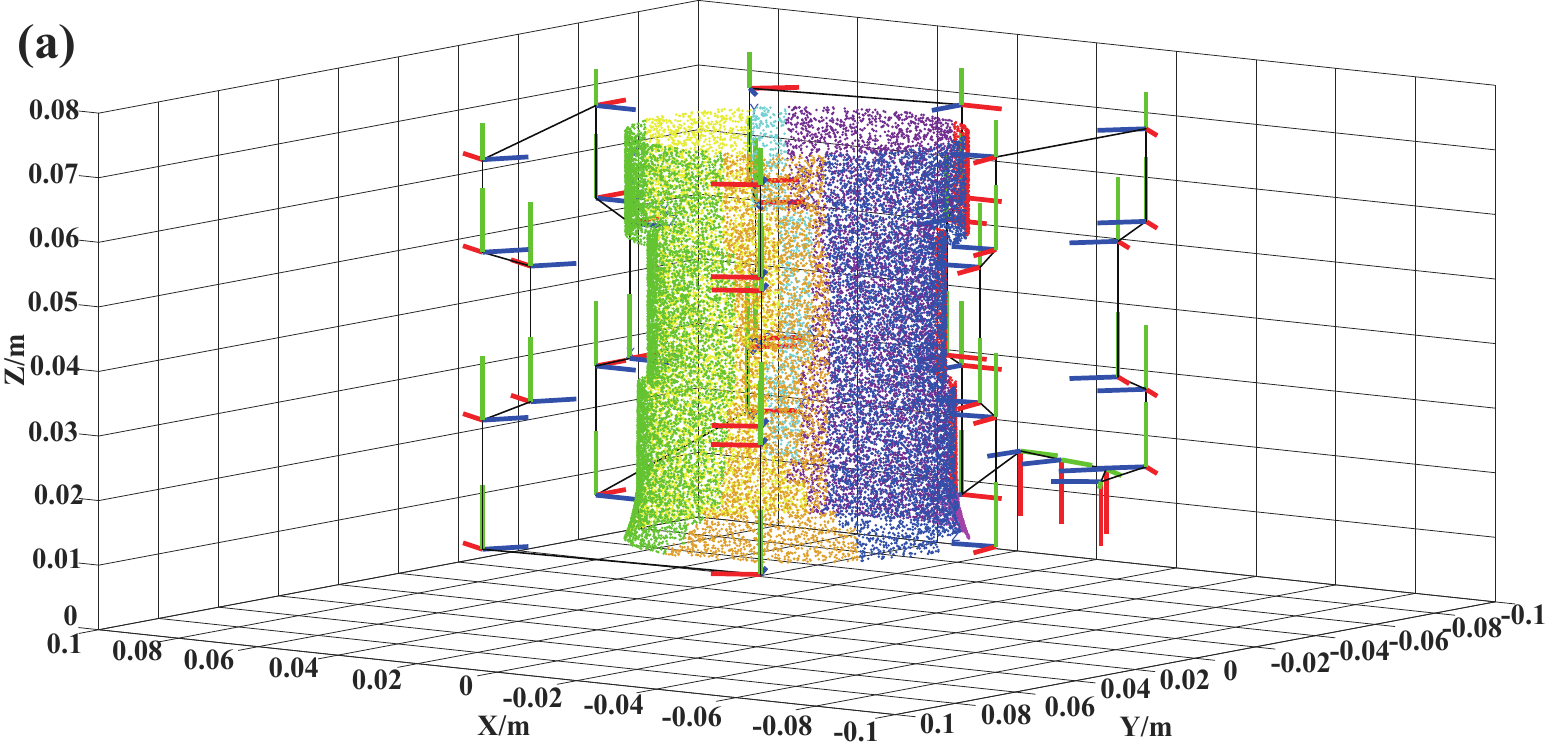}
        \label{fig: planning1}
    \end{subfigure}
    \begin{subfigure}
        \centering
        \includegraphics[scale=0.3]{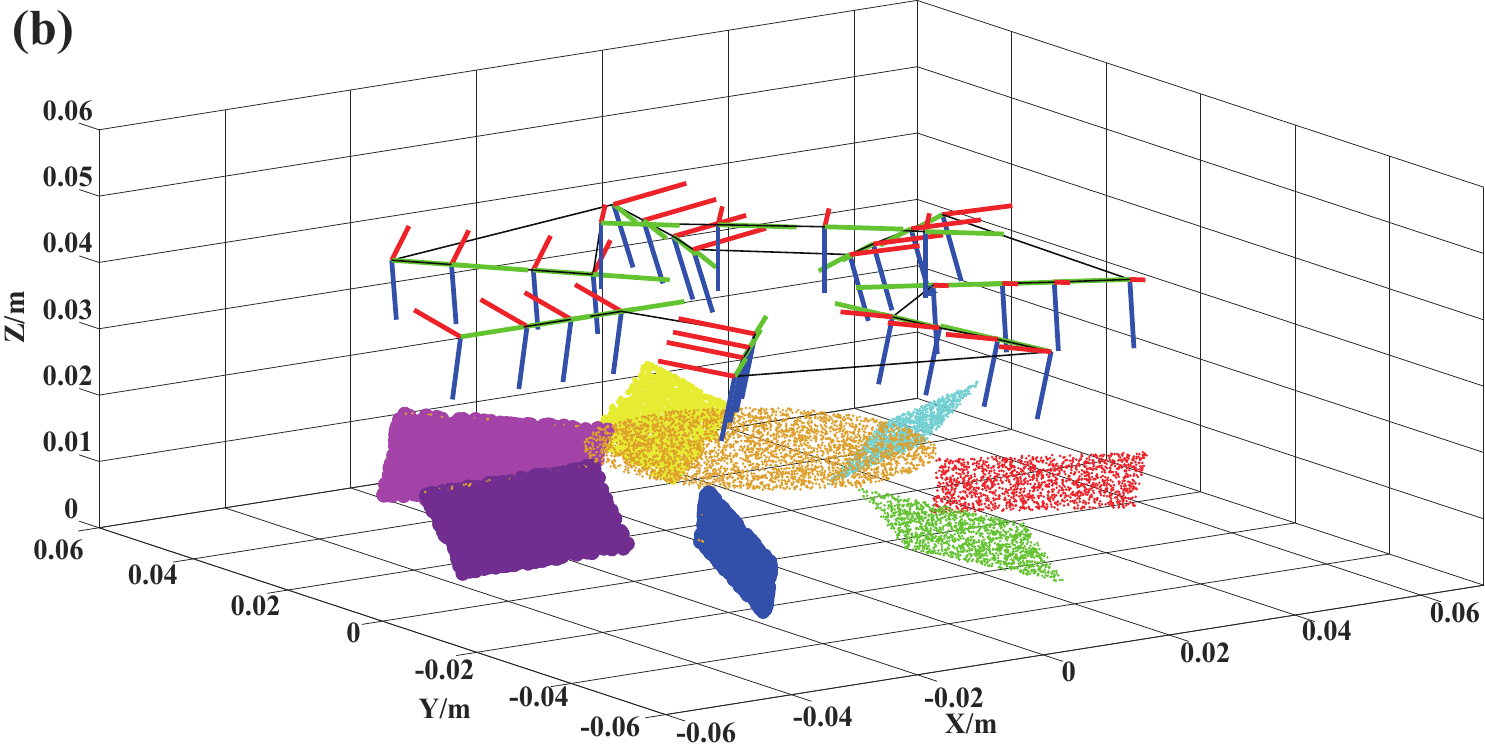}
        \label{fig: planning2}
    \end{subfigure}
     \begin{subfigure}
        \centering
        \includegraphics[scale=0.3]{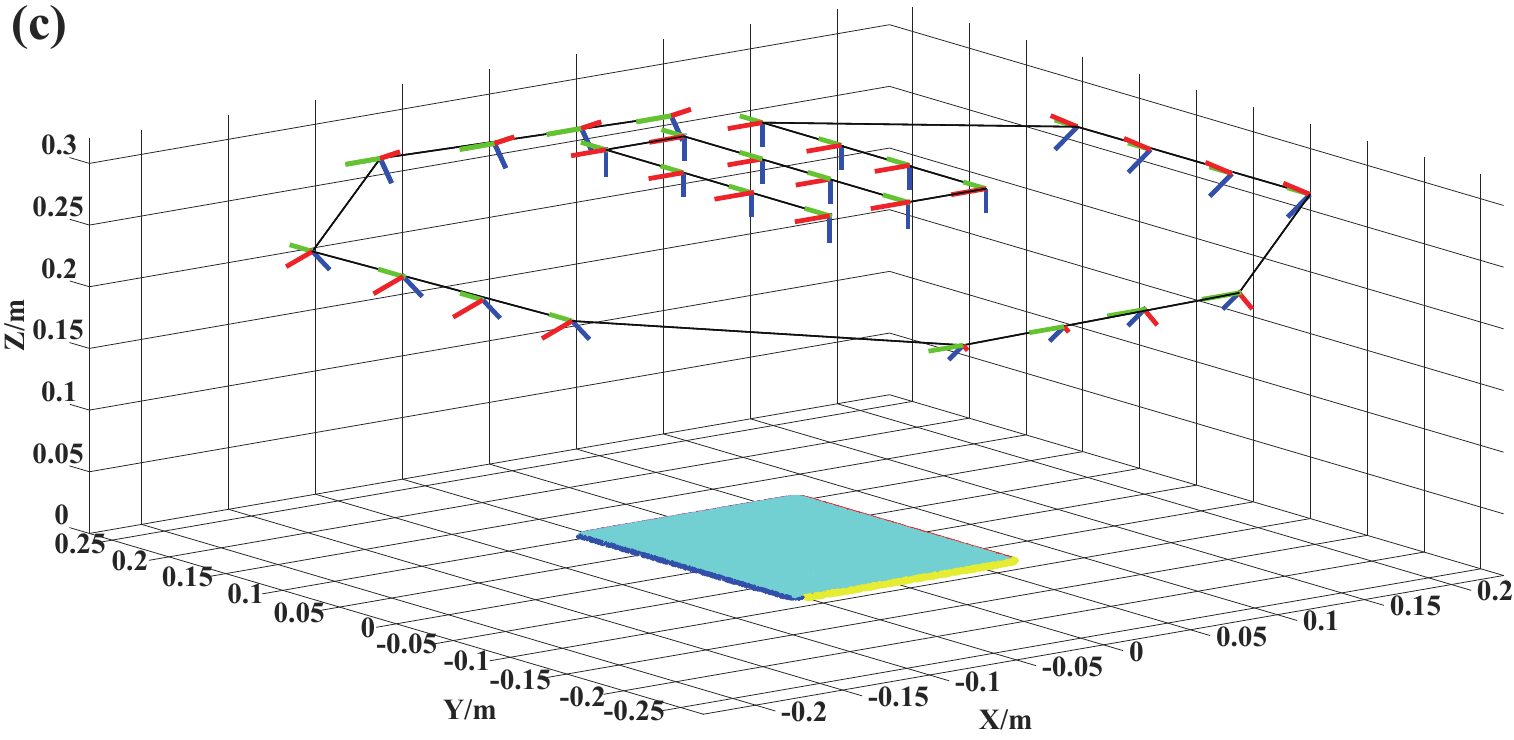}
        \label{fig: planning3}
    \end{subfigure} 
    \begin{subfigure}
        \centering
        \includegraphics[scale=0.3]{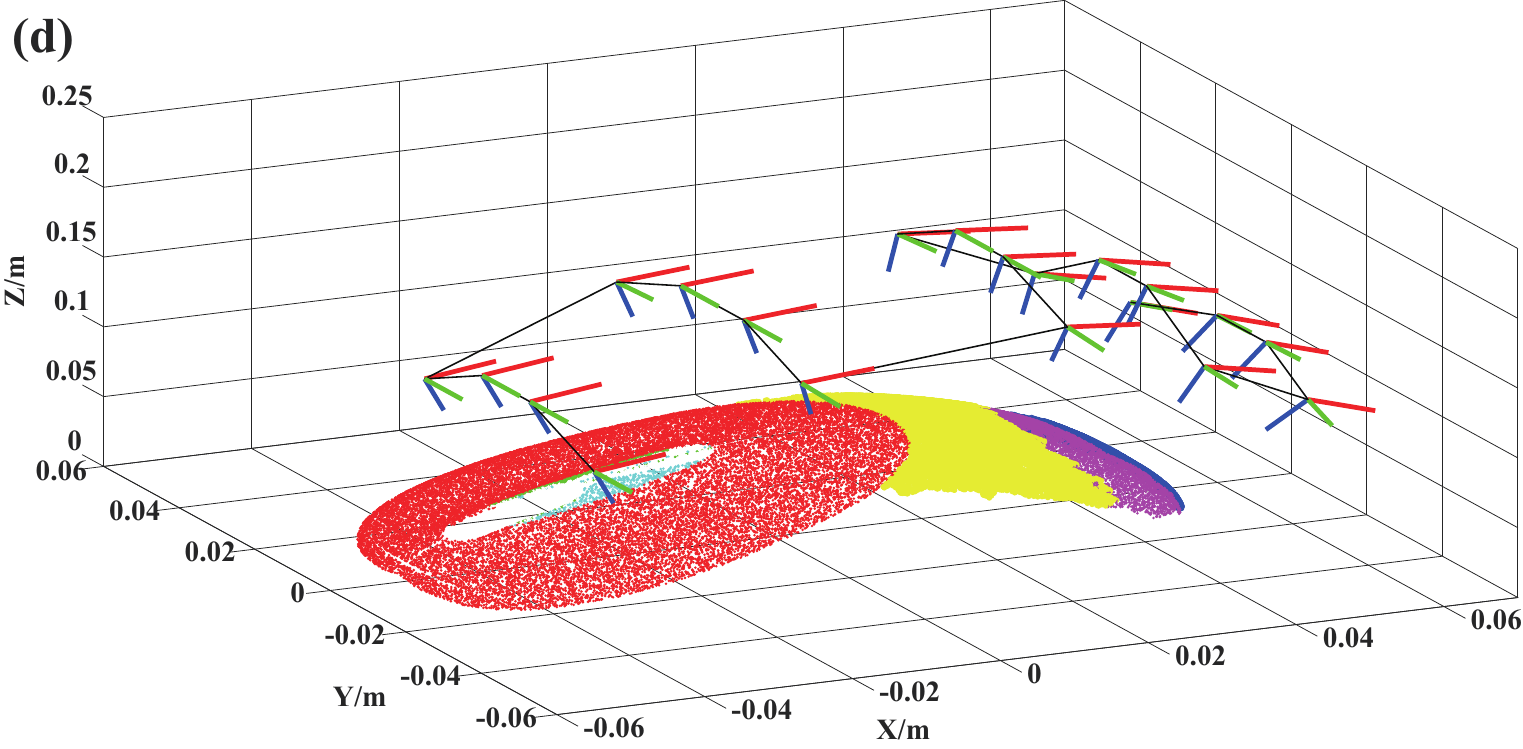}
        \label{fig: planning4}
    \end{subfigure}
    \caption{ Coverage path planning through the proposed
method, (a) Case 1: the camera lens; (b) Case 2: the computer fan; (c) Case 3: the tablet back cover; (d) Case 4: upper part of computer mouse }
	\label{fig: planning}
\end{figure}

With adaptive ROI-based path planning and PSO-based global path generation, a complete and near-optimal inspection path can be produced, which is visualized in Fig. \ref{fig: planning}. The number of viewpoints is 48, 48, 42, and 30 in Case 1-4 respectively, displayed by the frames. They show the pose of the robot's end-effector during the inspection task. The global path planning is demoted with a black line and every segmentated region has a corresponding local path. The different viewpoints are connected by straight lines in the optimal sequence. The robotics motion should follow this detection path to achieve full object coverage.  

\begin{figure}[h]
	\centering
	\includegraphics[scale=0.25] {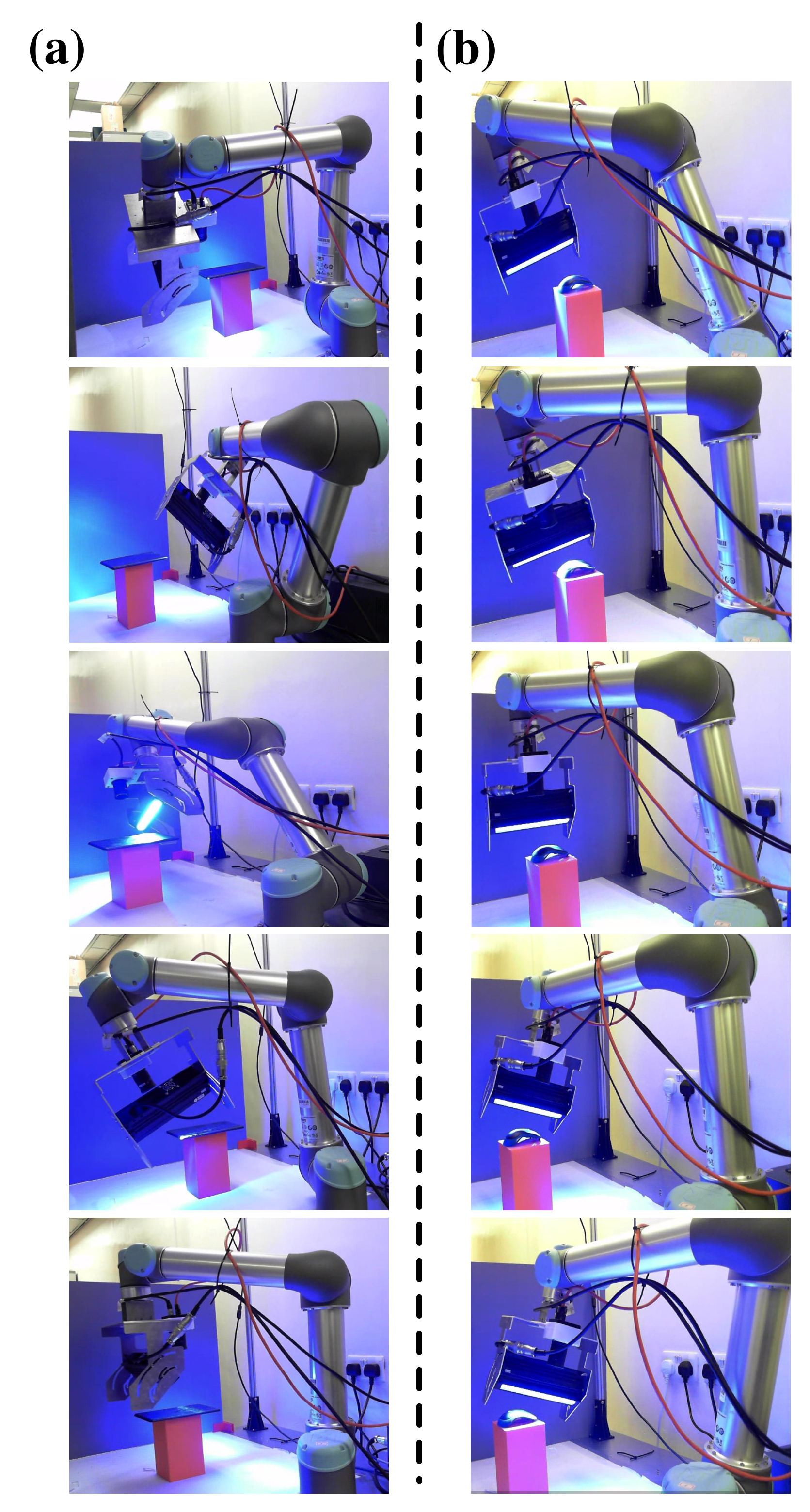}
	\caption{Experimental scenarios of 3C component surface inspection by the proposed method, (a) Case 3: the tablet back cover; (b) Case 4: upper part of computer mouse }
	\label{fig: Scenarios}
\end{figure}

\begin{figure}[h]
       \centering
       \includegraphics[scale=0.1] {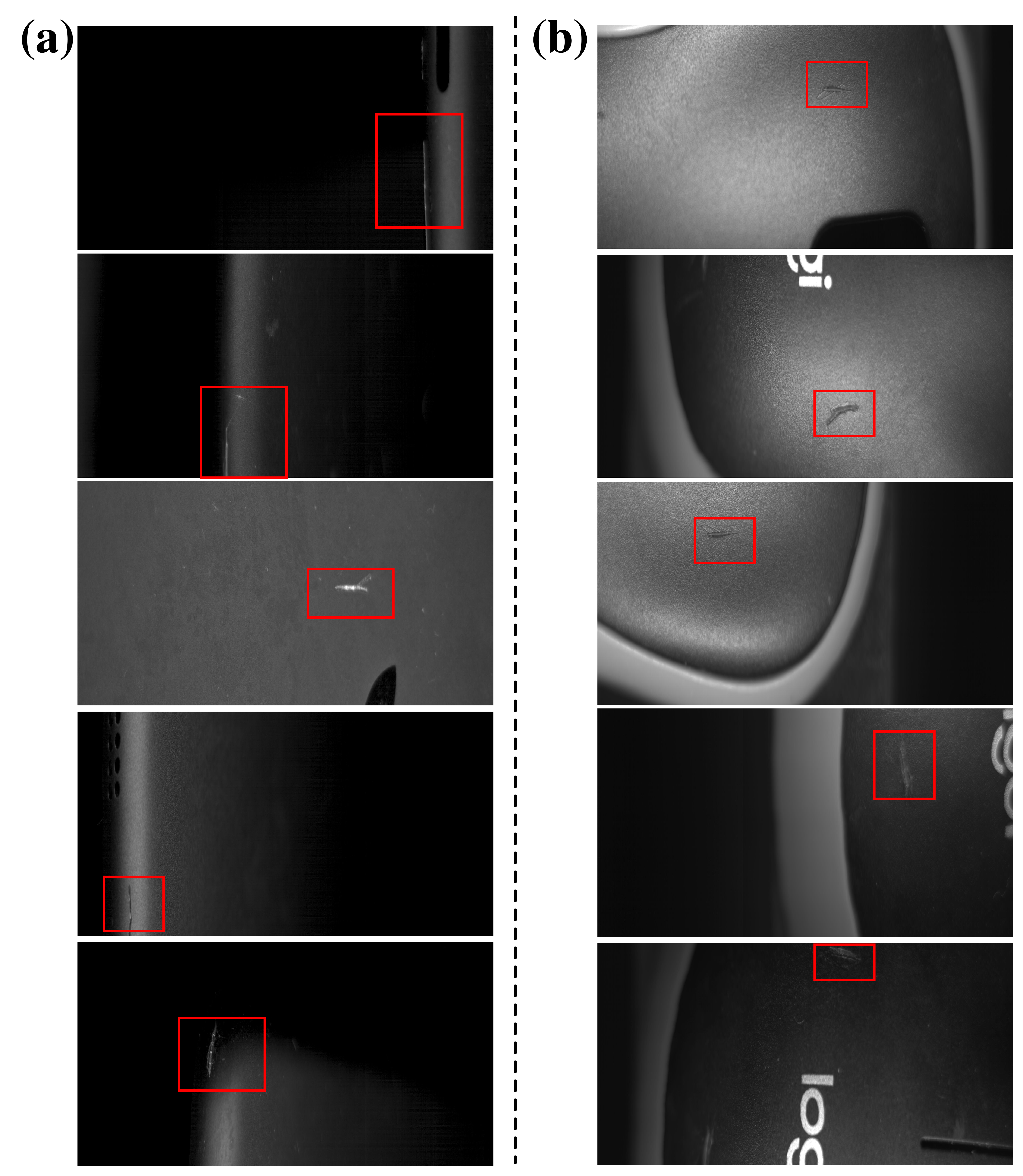}
       \caption{Representative defect images obtained by the proposed method, (a) Case 3: the tablet back cover; (b) Case 4: upper part of computer mouse}
	\label{fig: representative defect}
\end{figure}

We input the inspection paths to the automatic line scanning system to scan the tablet back cover and upper part of computer mouse in order to mimic the real defect inspection, as illustrated in Fig. \ref{fig: Scenarios}. 
Fig. \ref{fig: representative defect} illustrates the surface defects of these two objects. 
The segmentation process in our method results in uniformly geometric regions, allowing for the strategic selection of inspection viewpoints using an ROI-based technique congruent with line-scan camera parameters. This precision enables the system to detect surface defects with high clarity, surpassing human visual inspection capabilities, particularly in typically overlooked areas such as corners and curves.
The innovative aspect of our method lies in its integrated approach, combining advanced region segmentation with intelligent local and global path planning algorithms. This comprehensive approach facilitates meticulous inspection of surface imperfections, thereby contributing to enhanced processing and quality control within the 3C industry. 

\subsection{ Comparative analysis and verification}
\label{ssec:case_study} 

To further validate the proposed CPP method, two related cutting-edge scanning CPP methods, the clustering-based method TCluster~\cite{na2021cad} and the convex specular surface inspection method~\cite{huo2021sensor}, are applied as benchmark approaches for comparative analysis. The former is only used for viewpoint planning comparison, and the latter is applied to evaluate the actual defect inspection of the entire system.

In the local path definition, the region segmentation time was used as a measure of efficiency for region segmentation methods. Additionally, when defect results or coverage rates were similar, preference was given to the CPP method that generated fewer viewpoints as it was considered a more viable path planning approach~\cite{liu2020optimal}. The generation results are indicated in Table~\ref{tab:segmentation time} and Table~\ref{tab:Number of viewpoints} respectively. For region segmentation time, the proposed method used less time to finish this procedure. Due to the usage of RANSAC and more geometric features, the proposed method can obtain the subregions with planar/near-planar geometry efficiently. As for the viewpoints, our developed approach produces fewer viewpoints since more accurate region segmentation results and concise ROI generation. Conversely, the convex specular surface inspection method and TCluster method employed a more complex iteration process for viewpoint determination, as they struggled to segment objects precisely with intricate geometries.

\begin{figure}[!ht]
	\centering
	\includegraphics[scale=0.45] {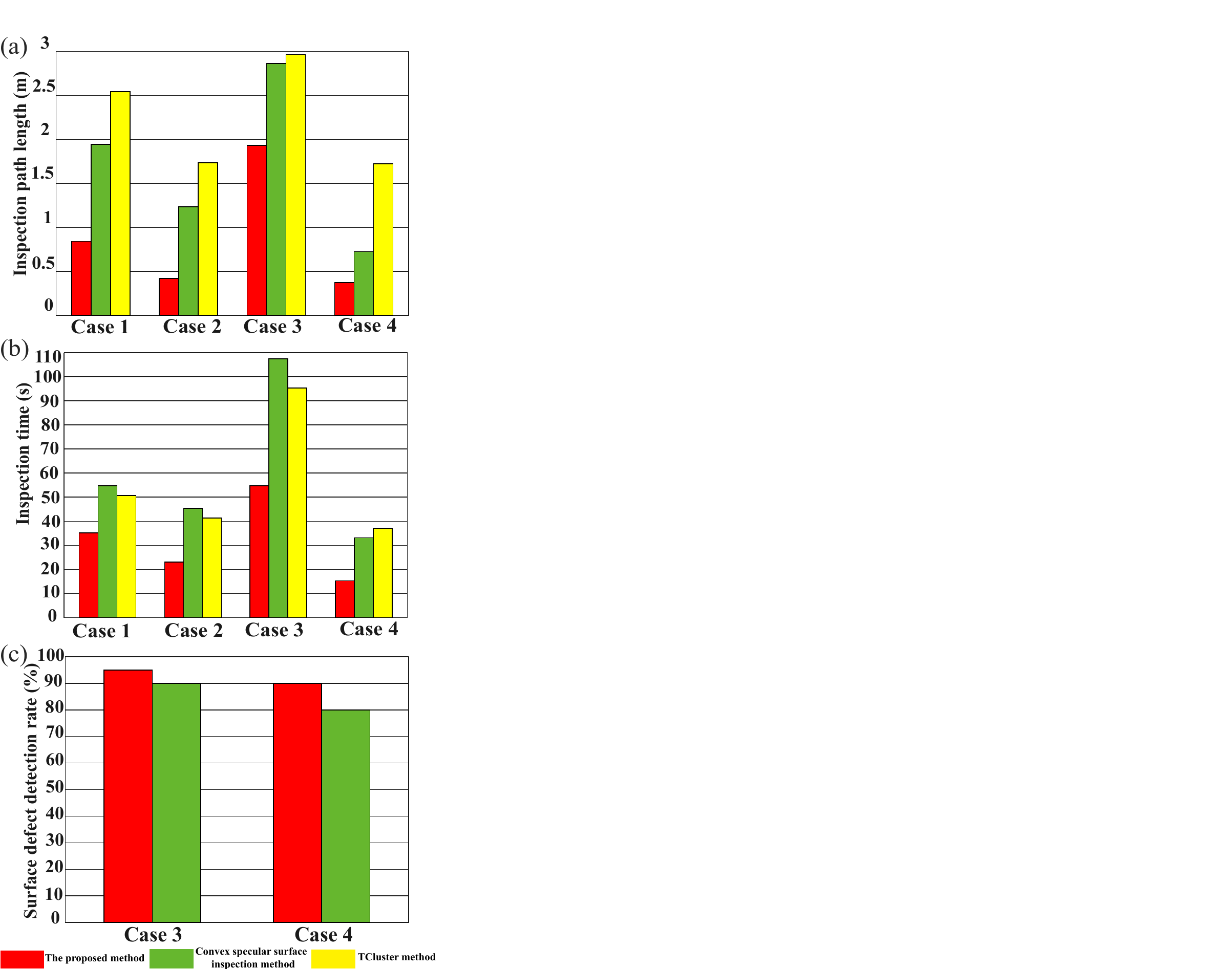}
	\caption{Results for global path planning and defect inspection of the proposed method, Convex specular surface inspection method, and TCluster method in Case 1-4. Comparison of (a) Inspection path length, (b) Inspection time, and (c) Surface defect detection rate}
	\label{fig: data analysis}
\end{figure}

For the global path planning, the comparison results are shown in Fig.~\ref{fig: data analysis}. The inspection path length and total detection time served as indicators of overall path efficiency in CPP methods. In this comparison, a defect detection method~\cite{huo2021sensor} is applied for surface quality inspection. Image enhancement is utilized to improve the quality and remove the noise based on the original captured images. Then, edge detection is critical to show sudden changes in gray-level play. Given the identified edges, a binary classifier is used to determine the approximated area of the edges. 
The surface defect detection rate serves as a direct measure of the actual effectiveness of defect acquisition, reflecting the accuracy of region segmentation and the quality of path planning.  Our proposed method demonstrates superior performance in terms of both inspection path length and time compared to benchmark approaches. While these benchmarks relied on relative optimization techniques, they were unable to produce a globally feasible inspection path for CPP. In contrast, our PSO-based approach adeptly tackles the TSP by setting realistic optimization targets and selecting feasible viewpoints that align with the inspection requirements. Although the performance of these two benchmark approaches is good, our method significantly enhances efficiency by completing inspections in less time and with shorter paths. This comprehensive evaluation positions our CPP method as a more effective alternative to conventional line scanning inspection techniques. The results underscore the value and practicality of our method in enhancing CPP for surface defect inspections.

    \begin{table}[h]
\centering
\caption{Comparison of region segmentation efficiency in Case 1-4}
\begin{tabular}{cccccc}

   \toprule
    Region segmentation time (s)&&  Case 1 &Case 2 &Case 3 &Case 4 \\
   \midrule
     \multicolumn{2}{c}{ The proposed method} & 8.06 & 2.19 & 10.31 & 3.34   \\
   \multicolumn{2}{c}{\begin{tabular}[c]{@{}c@{}}Convex specular surface\\ inspection method\end{tabular} } & 23.34&4.28&31.73&11.46   \\
  \multicolumn{2}{c}{ TCluster method}& 30.87&5.38&39.51&13.28\\
   \bottomrule
\end{tabular}
\label{tab:segmentation time}
\end{table}

    \begin{table}[h]
\centering
\caption{Comparison of number of viewpoints in Case 1-4}

\begin{tabular}{cccccc}

   \toprule
    Number of viewpoints&&  Case 1 &Case 2 &Case 3 &Case 4 \\
   \midrule
    
     \multicolumn{2}{c}{ The proposed method} &48&48&42&30    \\
     
   \multicolumn{2}{c}{\begin{tabular}[c]{@{}c@{}}Convex specular surface\\ inspection method\end{tabular} } & 96&66&78&54   \\
  \multicolumn{2}{c}{ TCluster method} &90&80 &75&63\\
   \bottomrule
\end{tabular} 
\label{tab:Number of viewpoints}
\end{table}

\section{Conclusion}    \label{sec:conclusion}
This paper proposes a systematic framework for an inspection CPP method for 3C component surfaces. According to this framework, a high-resolution line scanning sensor, mounted on a multi-DOF robotic manipulator, can execute surface scanning and detection precisely and flexibly, effectively addressing the line scanner viewpoint planning challenges for complex surface detection with accurate quality assessment requirements in the 3C industry. The developed methodology includes (1) a new hybrid region segmentation method based on the RANSAC and K-means clustering method; (2) an adaptive ROI method to define the local measurement paths; and (3) a PSO-based global optimization approach for the minimum inspection time. Four case studies verify the effectiveness and efficiency of this method. The results show it outperforms the state-of-the-art line scanning CPP method according to comparison. Overall, the proposed method can achieve precise and efficient surface inspection for 3C free-from components. It can be applied in the 3C industry and be extended to inspect other structures such as auto spare parts and industry-standard components. 

However, it should be noted that the proposed method may encounter challenges when applied to workpieces with complex structures, making it less suitable for parts with intricate shapes. Future research should focus on optimizing the design of the system end-effector to enhance the flexibility of the inspection framework. Additionally, exploring mathematical methods for optimal path planning and investigating the potential of information theory and deep learning techniques, such as convolutional neural networks, could further improve the effectiveness of the segmentation method.

\bibliographystyle{IEEEtran}
\bibliography{IEEEabrv, main}



%




\vfill

 


\end{document}